\documentclass[lettersize,journal]{IEEEtran}
\usepackage{amsmath,amsfonts}
\usepackage{algorithmic}
\usepackage{algorithm}
\usepackage{color}

\usepackage[dvipsnames]{xcolor}
\usepackage[Conny]{fncychap}
\usepackage{lipsum}

\usepackage{array}
\usepackage[caption=false,font=normalsize,labelfont=sf,textfont=sf]{subfig}
\usepackage{textcomp}
\usepackage{stfloats}
\usepackage{url}
\usepackage{verbatim}
\usepackage{graphicx}

\usepackage[noadjust]{cite}
\begin{document}

\title{Constrained Optimal Fuel Consumption of HEV:\\ A Constrained Reinforcement Learning Approach}

\author{Shuchang Yan,~
	             Futang Zhu,~
	             Jinlong Wu

\thanks{Shuchang Yan, Futang Zhu, and Jinlong Wu are with Product Planning and New Auto Technologies Research Institute, BYD Auto Industry Company Limited, Shenzhen, 518118, China (e-mail: ycyan@connect.hku.hk, zhu.futang@byd.com, wujl0519@163.com). 
	
This research was supported by the Department of Science and Technology of Guangdong Province (No. 2019ZT08L746). 
	}
}

\markboth{Journal of \LaTeX\ Class Files,~Vol.~14, No.~8, August~2021}%
{Shell \MakeLowercase{\textit{et al.}}: A Sample Article Using IEEEtran.cls for IEEE Journals}


\maketitle

\begin{abstract}
Hybrid electric vehicles (HEVs) are becoming increasingly popular because they can better combine the working characteristics of internal combustion engines and electric motors. However, the minimum fuel consumption of an HEV for a battery electrical balance case under a specific assembly condition and a specific speed curve still needs to be clarified in academia and industry. Regarding this problem, this work provides the mathematical expression of constrained optimal fuel consumption (COFC) from the perspective of constrained reinforcement learning (CRL) for the first time globally. Also, two mainstream approaches of CRL, constrained variational policy optimization (CVPO) and Lagrangian-based approaches, are utilized for the first time to obtain the vehicle's minimum fuel consumption under the battery electrical balance condition. We conduct case studies on the well-known Prius TOYOTA hybrid system (THS) under the NEDC condition; we give vital steps to implement CRL approaches and compare the performance between the CVPO and Lagrangian-based approaches. Our case study found that CVPO and Lagrangian-based approaches can obtain the lowest fuel consumption while maintaining the SOC balance constraint. The CVPO approach converges stably, but the Lagrangian-based approach can obtain the lowest fuel consumption at 3.95 L/100km, though with more significant oscillations. This result verifies the effectiveness of our proposed CRL approaches to the COFC problem.

\end{abstract}

\begin{IEEEkeywords}
HEV, COFC, constrained reinforcement learning, SOC balance constraint, CVPO, Lagrangian-based approach.
\end{IEEEkeywords}

\section{Introduction}
\IEEEPARstart{H}{EV} refers to a vehicle system that utilizes an internal combustion engine and a battery as a power source. It mainly comprises a control system, a drive system, a power auxiliary system, and a battery pack. It has the advantages of low fuel consumption and low emissions due to the cooperation of the engine, electric machine, and battery. HEV can be divided into micro hybrid, light hybrid, medium hybrid, and complete hybrid in terms of the dependence degree on electricity, into series type (extended range type), parallel type, and hybrid type based on the structure, and also into position 0 to position 4 (P0-P4) architecture concerning on motor position.

HEVs can track reference speed and fulfill other predetermined targets with the lowest energy consumption usually through three ways: one is to improve the energy conversion efficiency of each power train, one is the energy management control to utilize the energy smartly, and the last one is kinetic energy recovery to collect the energy produced by vehicle braking. Rule-based control (RBC) is widely used in energy control engineering practice. Fuel consumption under typical operating conditions such as WLTC, NEDC, or CLTC is one primary index that represents energy consumption level by careful energy control design. This aim is routinely achieved through software/hardware simulation before production and verified through drum tests or similar ones on vehicles after production. According to public information, BYD launched the first generation dual model HEV (ranked as DM 1.0, named as F3DM) with comprehensive operating conditions fuel consumption \( 2.7 \mathrm{~L} / 100 \mathrm{~km} \) in 2008, and caused a significant impact on vehicle market in China. In 2013 and 2018, DM 2.0 and DM 3.0 were released, respectively. In 2020, BYD launched the DM-i system, with which Qin PLUS Champion Edition's fuel consumption under WLTC can fall to \( 2.17 \mathrm{~L} / 100 \mathrm{~km} \).

Model predictive control (MPC)~\cite{rault1978model} and its application on vehicle energy control have constantly been developing in parallel and recently come to the stage in the industry (noted as predictive energy control, PEC). Its main content is that once a destination is selected,  PEC can minimize fuel consumption by adjusting the relevant powertrain output after knowing the future road conditions. In PEC research, reference~\cite{xie2019model} minimizes the sum of energy consumption cost and the battery life loss cost. The optimization is based on the plug-in HEV (PHEV) dynamics modeled by first-order derivative equation and battery dynamics modeled by first-order derivative equation. Reference~\cite{wang2020comparison} minimizes the sum of hydrogen consumption cost and the degradation cost of both the fuel cell stack and battery, respectively, by rule-based control and MPC. The optimization is based on the actual fuel cell/battery hybrid bus at the University of Delaware. 
Interested readers can refer to reference~\cite{lu2022hybrid} for a complete view of current PEC research and industry application status. PEC in the papers listed above must be appropriately simplified before it can be practically utilized. BMW, in 2012, applied PEC, which can adapt the gear to an appropriate position for the upcoming road condition~\cite{BWMnews}. In 2022, the Nezha S range-extended version benefiting from PEC is claimed that the fuel-saving rate is increased by   3\%, and the pure electric cruising range is increased by 2\%~\cite{netaautoNETAHome}. In 2023, Geely Galaxy L6's PEC is based on driver/user habits, and the energy saving rate is announced amazingly 15\%~\cite{galaxy}.

Whether using RBC or PEC, as engineers, we all want to know what the optimal (lowest) fuel consumption is for a specific assembly condition and reference speed trajectory. Also, the national standard's fuel consumption evaluation of HEVs needs to be based on battery electricity balance condition, that is, the starting and ending SOC of the battery under specific working conditions should be kept as close as possible~\cite{interregsInterRegsGBT}, In this work, we want to address the burning question:

\vskip 0.1 cm
\textit{How do we get the lowest fuel consumption for a vehicle dynamic model and a reference speed trajectory while keeping the battery SOC balance?}
\vskip 0.1 cm

We named this problem the COFC problem. Mathematically speaking, this is an optimal control problem, and a proper optimization approach should be utilized.

RBC or PEC are myopic and look-ahead optimization, respectively, and are known to be local optimization approaches. Heuristic methods like genetic algorithm~\cite{holland1992genetic} are also local optimization approaches, and their optimality, stability, and feasibility can not be guaranteed. Dynamic programming (controller) is a well-known global optimization approach but suffers from the curse of dimensionality, and this will cause optimal control of a physical system with multiple inputs, states, or a long control horizon to be intractable. On the other hand, it is difficult to deal with constraints on states for dynamic programming. Later, reinforcement learning (RL, also called approximate dynamic programming in the operation research community) is developed to handle this optimal control problem for the general nonlinear dynamic system. The basic idea of RL is to learn the Bellman equation approximated by the obtained states and rewards after putting actions into the controlled system (called the environment in RL). Typical RL utilizes value iteration usually for discrete decisions (e.g., DQN~\cite{mnih2013playing}), policy iteration usually for continuous decisions (e.g., DDPG~\cite{lillicrap2015continuous}), or their combined approach (e.g., PA-DDPG~\cite{xiong2018parametrized}) to gradually approximate the Bellman equation.

Few works applying RL to the COFC problem of HEV or EV are reported. Recent reference~\cite{lian2020rule} utilizes DDPG and DQN to investigate the THS system's optimal fuel consumption  (OFC). Reference~\cite{liu2023safe} utilizes the THS system and the penalty function methods to transform a constrained optimization problem into an unconstrained one. Reference~\cite{wu2024confidence} considers the storage and fuel cell dynamics of a fuel cell electric vehicle, and the penalty method is utilized for the SOC deviation from the target SOC. Penalty method utilized for constraints can also be seen at~\cite{liu2021intelligent, wu2020battery}. However, the literature listed above has a similar problem: The penalty method dealing with constraints can not ensure feasibility at most times and usually gives a compromise solution (as a trade-off between optimality and feasibility for the OFC problem). The current research work we surveyed all focuses on the OFC problem rather than the COFC problem. The constrained reinforcement learning (CRL) method must be introduced and developed to solve the COFC problem.

The research on CRL grows in two lines: one is the Lagrangian-based approach, and the other is to transfer the RL as a probabilistic inference problem. Both approaches aim to make the optimization problem feasible within the constraints. The Lagrangian-based approach transfers a constrained problem into an unconstrained one solved by primal-dual variable iteration~\cite{chow2018risk, liang2018accelerated}. Because of the imbalanced learning rate of the primal and dual problems, the primal-dual iteration often suffers the problem of numerical instability and lack of stability guarantee~\cite{chow2018lyapunov}. To simplify the primal-dual iteration, low-order Taylor expansion for approximating the constrained optimization is introduced to solve the dual variable~\cite{achiam2017constrained}. However, the approximation error may cause the constraint to be satisfactory. The other emerging line is to treat the constrained problem as a probabilistic inference problem, named CVPO, first publicly released in 2022~\cite{liu2022constrained}. Its procedure has two steps: the expectation step (E-step) for seeking an optimal and feasible non-parametric variational distribution analytically and the maximization step (M-step) for parameterized policy training by a supervised learning procedure. This algorithm design ensures that the E step has an optimality guarantee and the M step has a worst-case constraint violation bound and robustness guarantee against worst-case training iterations. Unfortunately, we have not seen any application of CRL (both the Lagrangian-based and CVPO approaches) in the vehicle research domain until now.




In this work, we propose the COFC problem in this section and give its mathematical model in section II, introduce the CVPO approach and Lagrangian-based approaches, including  deep deterministic policy gradient (lr-DDPG) and soft actor-critic (lr-SAC)~\cite{haarnoja2018soft}
of off-policy learning, and first-order constrained optimization in policy space (lr-FOCOPS)~\cite{zhang2020first}, proximal policy optimization (lr-PPO)~\cite{schulman2017proximal}, trust region policy optimization (lr-TRPO)~\cite{schulman2015trust} and constrained policy optimization (lr-CPO)~\cite{achiam2017constrained} of on-policy learning in Section III, and results and the followed analysis are given in Section IV, finally ends up with conclusions and discussions on the COFC-related open questions in Section V and Section VI respectively. The main contributions of this work are as follows,


\par(i) This is the first work that proposes the COFC problem and gives its mathematical description worldwide from the perspective of the CRL. The COFC problem is essential as its answer gives a solid hint for evaluating the vehicle fuel consumption level, which is urgently needed for PEC design among almost all auto companies.

\par(ii) This is the first work introducing the CRL, including the CVPO and Lagrangian-based approaches to tackle the COFC problem in the vehicle research domain. The introduction of CRL, especially the CVPO approach, truly moves one essential step through the feasible, stable, and optimal optimization, and this privilege definitely can not be compared to the existing penalty method for the OFC problem.


\par(iii) To the best of our knowledge, this is the first work in which a detailed framework build-up procedure and follow-up analysis is given for academic researchers and engineers with almost complete flexibility to analyze the vital decisions on the fuel consumption of a particular vehicle in optimal cases. Our framework can also analyze and calculate other indices of vehicles, such as acceleration time.


\section{Problem Description}

\subsection{Preliminary Work}

We consider a general nonlinear dynamic system in a discrete form as follows:
\begin{gather}
	s_{t+1}=F\left(s_{t}, a_{t}\right)\label{d1} \\
	y_{t+1}=Z\left(s_{t}, a_{t}\right)\label{d2}
\end{gather}
Where \( s_{t} \) is the state vector of the dynamic system at time \( t, F \) is the transition function of the dynamic system, \( a_{t} \) is the applied control to the dynamic system at time \( k,  y_{t+1} \) is the observation vector of the dynamic system at time \( t \), and \( Z \) is observation function of the dynamic system.

One task in optimal control is to design a control rule to manipulate (all or part of) states and observations to the desired value at a specified time. In the RL community, the control rule to make control input is also named policy, represented as \( \pi_{\theta}=\pi(a|s; \theta) \) with parameter vector \( \theta \). In each step, we obtain transition ($s_{t}$, $a_{t}$, $s_{t+1}$) of dynamic system (\ref{d1})-(\ref{d2}), and further define the reward as \( r_{t}=R\left(s_{t}, a_{t}\right) \) and cost as \( c_{t}=C\left(s_{t}, a_{t}\right) \). Then we add the termination signal (noted as {\sffamily  done}) and task status (noted as {\sffamily  False} and {\sffamily  True}, meaning the dynamic system is ongoing and over, respectively). In each step transition we obtain a six-element  tuple ($s_{t}$, $a_{t}$,  $s_{t+1}$,  $r_{t}$,  $c_{t}$, {\sffamily done}) . For one trajectory \( \tau=\left\{s_{0}, a_{0}, s_{1}, \ldots\right\} \) under policy \( \pi_{\theta}=\pi(a|s ; \theta) \), we have the discounted expect return of reward \( J_{r}(\pi)=E_{\tau \sim \pi}\left[\sum_{t=0}^{\infty} \gamma^{t} r_{t}\right] \), the discounted expectation of return of cost \( J_{c}(\pi)=E_{\tau \sim \pi}\left[\sum_{t=0}^{\infty} \gamma^{t} c_{t}\right] \) and $\gamma$ is the discount factor.

For our problem, the dynamic system is the vehicle dynamic model, and we need to design a control rule, more specifically, a CRL controller, to investigate the lowest fuel consumption within the SOC balance constraint.

\subsection{Problem Description}

Given a speed trajectory and a vehicle dynamic model, our task is \textit{to implement the CRL approaches and their decisions to control the vehicle dynamics to follow the given reference speed with the SOC balance constraint.} The mathematical description of the COFC problem is as follows,
\begin{gather}
\pi^{*}=\arg \max _{\pi} J_{r}(\pi)  \label{o1}
\end{gather}
\vskip  -0.3cm
~~~~~~~~~~~~~ s.t.
\begin{align}
J_{c}(\pi)  &\leq \varepsilon_{1} \label{o2}\\
 s_{t+1} &= F\left(s_{t}, a_{t}\right) \label{o3}
\end{align}
Where \( \pi_{\theta}=\pi(a|s; \theta) \) is the CRL controller we need to design, and \( \varepsilon_{1} \) is the maximum allowed discounted expectation of return of cost for one trajectory. The vehicle dynamic system is represented by (\ref{o3}). In the following paper, we may ignore the subscript representing the time and will make a unique statement if there is any possibility of ambiguity.

\subsection{SOC Constraint}

\begin{figure}[!tbh]
	\centering
	\includegraphics[width=3.2in]{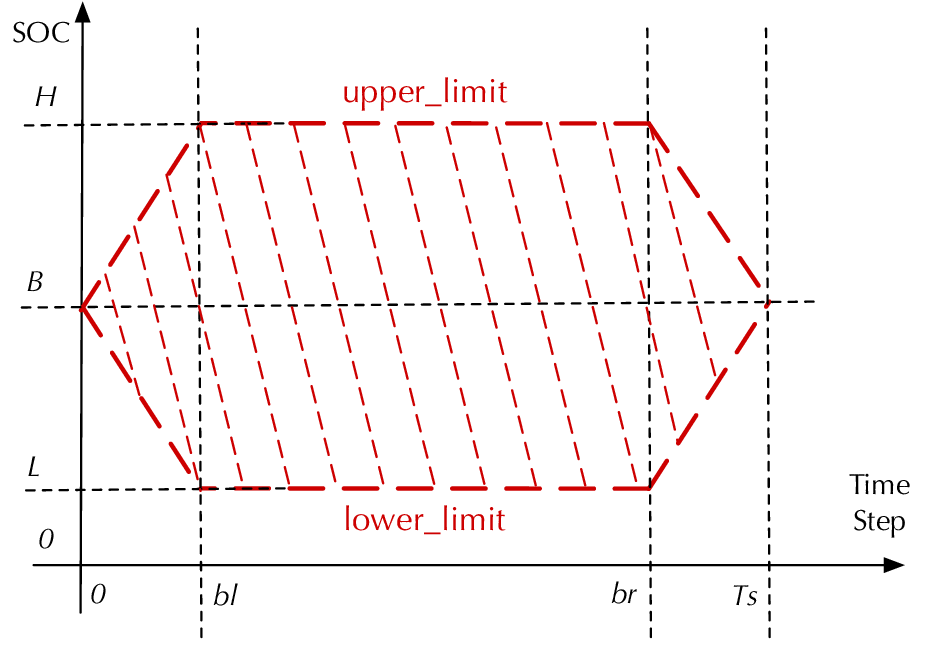}
	\caption{Battery SOC-allowed range (filled with red dotted lines).}
	\label{mm4}
\end{figure}

Based on national standards and engineering practices, we need the SOC of the battery around or better fixed at the same value (called the balance point in industry) at the beginning and end for a trajectory. This value of the balance point may differ slightly, and we adopt the SOC balance point at \( 50 \% \) here. The SOC upper and lower limits are shown in Fig.\ref{mm4}.

The mathematical description for the range filled in red is as follows,

\begin{align}
	\text{upper\_limit}=\left\{
	\begin{array}{rcl}
		\frac{(H-B)} {bl}  \times t +B ,&& {t      \leq   bl}\\ \\
		\frac{(H-B)} {br-Ts}  \times (t-Ts) + B,&&{t     >  br} \\ \\
		H,&&{\text{others}}
	\end{array} \right. 
\end{align}

\begin{align}
\text{lower\_limit}=\left\{
\begin{array}{rcl}
	\frac{(L-B)} {bl}  \times t  + B,&&{t      \leq    bl}\\ \\
	\frac{(L-B)} {br-Ts}  \times (t-Ts) + B,&&{t      >     br}\\ \\
	L, & &{\text{others}}
\end{array} \right. 
\end{align}


\noindent Where  $\text{upper\_limit}$  and $\text{lower\_limit}$  are the upper and lower limit lines of the SOC-allowed range, respectively. The $H$, $L$, and $B$ are the highest SOC value, lowest SOC value, and initial SOC value, respectively. The dynamic system starts from the time step $0$  and ends at the time step $Ts$. $bl$ and $br$ are the time steps at which the SOC limit trend changes. Note that the SOC-allowed range can be set as needed. In this work, we set the cost as 
\begin{align}
\text { cost }=\max (\text {SOC}- \text {upper\_limit}, 0) \notag\\
+\max (\text {lower\_limit}- \text {SOC}, 0)
\end{align}

\section{Algorithm Design}
Here, we introduce the CRL approaches, including Lagrangian-based and CVPO approaches. Since the primary goal of this work is to solve the COFC problem, we will list the principles and critical processes of the approaches. For particular content, we recommend that interested readers carefully read the literature we list and the data we put on the website or further discuss it via email if needed.

\subsection{Lagrangian-based Approaches}

The COFC problem is written in a Lagrangian form as follows,
\begin{gather}
	\left(\pi^{*}, \lambda^{*}\right)=\arg \min_{\lambda \geq 0} \max _{\pi} J_{r}(\pi)-\lambda\left(J_{c}(\pi)-\epsilon_{1}\right)\label{nn1}
\end{gather}
Where $\lambda$  and $\lambda^{*}$ are the Lagrange multiplier (or dual variable) and optimal  Lagrange multiplier (or optimal dual variable) respectively, and $\pi^{*}$ is the optimal policy, in our context, the optimal CRL controller.

\begin{figure}[!tbh]
	\centering
	\includegraphics[width=3.2in]{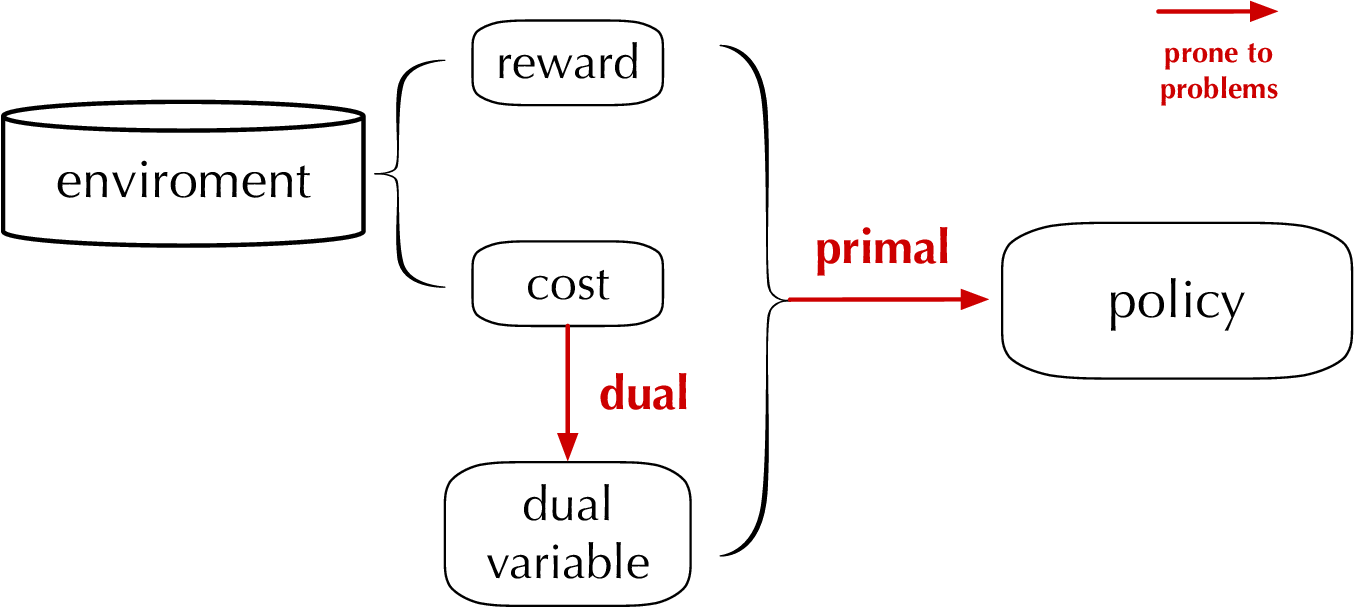}
	\caption{The perspective of Lagrangian-based approaches.}
	\label{li}
\end{figure}

\textit{Iteratively solving the $\text{min}$ (w.r.t $\lambda$) -- $\text{max}$ (w.r.t $\pi $), corresponding to the dual and primal iteration respectively, is the primary process of the Lagrangian-based approaches.} Noted the dual variable $\lambda$ is $0$ if  $J_{c}(\pi)<\epsilon_{1}$ and $\lambda$ is $+\infty$ if  $J_{c}(\pi)>\epsilon_{1}$.  Such an extensive range of  $\lambda$ will cause the possibly inaccurate approximation of $\lambda$ in dual process and possibly unstable policy gradient calculation in primal process, which is prone to problems. The consideration above is shown in Fig.\ref{li} as follows,

In this work, we adopt the latest and more robust PID Lagrangian-based approaches as in~\cite{stooke2020responsive}.  
The critical steps of the original Lagrangian approach are in {\bf  Algorithm 1}.

\begin{algorithm}[htbp]
	\caption{The Lagrangian approach.} 
	1:\hspace*{0.02in} {\bf procedure:} CRL \( \left(\pi_{\theta_{0}}(\cdot \mid s)\right) \)\\
	2:\hspace*{0.02in}  \quad \( \pi_{\theta_{0}}(\cdot \mid s) \) initialization\\
	3:\hspace*{0.02in}   \quad  \( J_{c} \leftarrow\{\} \quad \triangleright \) collect\\
	4:\hspace*{0.02in} \quad   {\bf repeat}\\
	5: \hspace*{0.06in} \quad  Sample environment: \( \quad \triangleright \) a minibatch\\
  	6:\hspace*{0.25in}  \quad \quad  $a \hspace{1mm}\sim \pi(\cdot\mid s ; \theta)$\\ 
	7:\hspace*{0.25in}  \quad \quad  $s^{\prime}\sim F(s, a) $\\
	8:\hspace*{0.25in}  \quad \quad	 $ r ~\sim R(s, a)$\\
	9:\hspace*{0.25in}  \quad \quad $ c ~\sim C(s, a) $\\
	10: \hspace*{0.01in} \quad  Apply feedback control:\\
	11: \hspace*{0.25in} \quad  Store sample estimate \( \hat{J}_{c} \) into \( J_{c} \)\\
	12: \hspace*{0.25in} \quad  Calculate dual vairiable $\lambda$ of min problem in ({\ref{nn1}}) \\
	13: \hspace*{-0.02in} \quad   Update $\pi(\cdot \mid s ; \theta)$  of  max problem  in ({\ref{nn1}}) through \hspace*{1.02cm} policy gradient  \\
	15:\hspace*{-0.1cm}  \quad   {\bf until} converged \\
	16:\hspace*{-0.1cm}   \quad {\bf return}~\( \left(\pi_{\theta_\text{final}}(\cdot \mid s)\right) \)\\
	17:\hspace*{0.02in} {\bf end procedure} 
\end{algorithm}

In {\bf  Algorithm 1}, steps 11 and  12 are the dual process, and 
step 13 is the primal process, $s$ and $s^{\prime}$ are the current state and next state respectively, {\textit R} and {\textit C} are the reward and cost functions respectively. Then, PID control is employed to update the dual variable $\lambda$ to become the PID Lagrangian approach. In {\bf  Algorithm 2}, the function of the derivative term is to act against cost increases but does not impede decreases as the mapping to a non-negative interval. The function of the integral term is to ensure zero steady-state error. If we do not consider the proportional and derivative terms, {\bf Algorithm 2} will revert to the traditional Lagrangian-based approach.

\begin{algorithm}[htbp]
	\caption{Dual variable update by PID control.} 
	1:\hspace*{0.02in} Set PID parameters:  \( K_{P}, K_{I}, K_{D} \geq 0 \)\\
	2:\hspace*{0.02in}  Integral: \( I \leftarrow 0 \)\\
	3:\hspace*{0.02in}   Previous cost: \( J_{c, prev} \leftarrow 0 \)\\
	4:\hspace*{0.02in} {\bf repeat} at each iteration \( k \)\\
	5: \hspace*{0.20in} Receive cost \( J_{c} \)\\
	6:\hspace*{0.25in}  \( \Delta \leftarrow J_{c}-\varepsilon_{1} \)\\ 
	7:\hspace*{0.25in} \( \partial \hspace{0.9mm} \leftarrow\left(J_{c}-J_{c, \text { prev }}\right)_{+}\)\\
	8:\hspace*{0.25in}  \( I\hspace{1.1mm}\leftarrow(I+\Delta)_{+} \)\\
	9:\hspace*{0.25in}  \( \lambda \hspace{1mm}\leftarrow\left(K_{P} \Delta+K_{I} I+K_{D} \partial\right)_{+} \)\\
	10: \hspace*{0.12in} \( J_{c, \text { prev }} \leftarrow J_{c} \)\\
	11: \hspace*{0.20in}{\bf return} \( \lambda \)
\end{algorithm}

 \subsection{CVPO Approach}
 
The CVPO approach insightfully views the COFC problem as a probabilistic inference problem, inferring actions $a$  with high rewards but within cost limit $\epsilon_{1}$.

 \textit {Basic Idea and Mathematical Description}:
The \( p_{\pi}(\tau) \) represents the probability of a trajectory \( \tau \) under the policy \( \pi \).
Optimality variable O represents the event of maximizing the reward, and it has the property:
\begin{gather}
 p(O=1 \mid \tau) \propto \exp \left(\sum_{t} \gamma^{t} r_{t} / \alpha\right) 
 \end{gather}
  Where  \( \alpha  \) is the temperature parameter. 
  Then the lower bound of the log-likelihood of optimality under the policy \( \pi \) is:
 \begin{gather}
 	\log p_{\pi}(O=1)=\log \int p(O=1 | \tau) p_{\pi}(\tau) d \tau \notag \\
 	\geq \mathbb{E}_{\tau \sim q}\left[\sum_{t=0}^{\infty} \gamma^{t} r_{t}\right]-\alpha D_{\mathrm{KL}}\left(q(\tau) \| p_{\pi}(\tau)\right)=\mathcal{J}(q, \pi)
 \end{gather}
Where \( q(\tau) \) is an auxiliary trajectory distribution and \( \mathcal{J}(q, \pi) \) is the evidence lower bound (ELBO).  For KL divergence $D_{\mathrm{KL}}(\cdot \| \cdot)$, unfamiliar readers are recommended to refer to~\cite{fox2012tutorial}.

We limit the choices of \( q(\tau) \) within a feasible distribution family subject to constraints and define the feasible distribution family regarding the threshold \( \epsilon_{1} \) as
 \begin{gather}
 \Pi_{\mathcal{Q}}^{\epsilon_{1}}:=\{q(a|s): \mathbb{E}_{\tau \sim q}\left[\sum_{t=0}^{\infty} \gamma^{t} c_{t}\right]<\epsilon_{1}, a \in \mathcal{A}, s \in \mathcal{S}\}
  \end{gather}
Where $\mathcal{Q}$   represents that the $  \Pi_{\mathcal{Q}}^{\epsilon_{1}}$ is related to  $q$,  $\mathcal{A}$ and $\mathcal{S}$ represent the action space and state space respectively. Afterwards, by factorizing the trajectory distributions:
 \begin{align}
 	\small
 	q(\tau)&=p\left(s_{0}\right) \prod_{t \geq 0} p\left(s_{t+1}| s_{t}, a_{t}\right) q\left(a_{t} | s_{t}\right), \forall q \in \Pi_{\mathcal{Q}}^{\epsilon_{1}}\\
 	p_{\pi_{\theta}}(\tau)&=p\left(s_{0}\right) \prod_{t \geq 0} p\left(s_{t+1} | s_{t}, a_{t}\right) \pi_{\theta}\left(a_{t}| s_{t}\right) p(\theta)
 \end{align}
 Where \( \theta \in \Theta \) is the policy parameters, \( \Theta \)  represents parameter space, and \( p(\theta) \) is a prior distribution.  By canceling the transitions, we obtain the ELBO as follows:
 \begin{gather}
 	\small
 \hspace{0mm}	\mathcal{J}(q, \theta)= 
 	 \mathbb{E}_{\tau \sim q}\left[\sum_{t=0}^{\infty}\left(\gamma^{t} r_{t}-\alpha D_{\mathrm{KL}}\left(q\left(\cdot | s_{t}\right) \| \pi_{\theta}\left(\cdot \mid s_{t}\right)\right)\right)\right] \notag\\
 	+\log p(\theta), \quad \forall q\left(a|s_{t}\right) \in \Pi_{\mathcal{Q}}^{\epsilon_{1}}\label{elbo}
 \end{gather}

\begin{figure}[!tbh]
 	\centering
 	\includegraphics[width=3.5in]{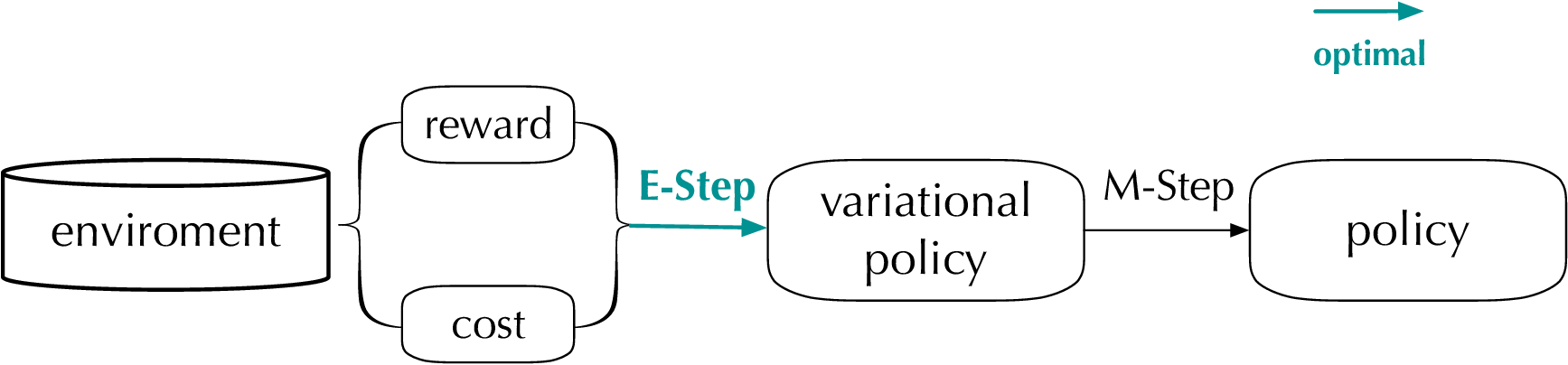}
 	\caption{The perspective of the CVPO approach.}
 	\label{cv}
 \end{figure}

\textit {Iteratively at E-step optimizing the new lower bound \( \mathcal{J}(q, \theta) \) w.r.t \( q \) within \( \Pi_{\mathcal{Q}}^{\epsilon_{1}} \)  and at M-step optimizing \( \pi \)  w.r.t \( \theta \) within \( \Theta \)  through an Expectation Maximization (EM) fashion is the primary process of the CVPO approach as in Fig.\ref{cv}.}

 \subsubsection{E-Step} The goal  is to obtain \( q \)   by maximizing	$\mathcal{J}(q, \theta)$ after fixing  the  $\theta$ = $\theta_{i}$  at iteration $i$ as follows, 
 \begin{gather}
 	\overline{\mathcal{J}}(q)=\mathbb{E}_{\rho_{q}}\left[\mathbb{E}_{q(\cdot \mid s)}\left[Q_{r}^{\pi_{\theta_{i}}}(s, a)\right]-\alpha D_{\mathrm{KL}}\left(q \| \pi_{\theta_{i}}\right)\right] \\[9pt]
 	\text {s.t. } \quad \mathbb{E}_{\rho_{q}}\left[\mathbb{E}_{q(\cdot \mid s)}\left[Q_{c}^{\pi_{\theta_{i}}}(s, a)\right]\right] \leq \epsilon_{1} \label{qc}
 \end{gather}
 Where  \( \rho_{q}(s) \) is the stationary state distribution induced by \( q(a|s) \) and \( \rho_{0} \).  \( Q_{r}^{q}(s, a) \) and  \( Q_{c}^{q}(s, a) \)  are the discounted return of reward and cost, respectively, estimated by: 
\begin{align}
Q_{r}^{q}(s, a)=Q_{r}^{\pi_{\theta_{i}}}(s, a)=\mathbb{E}_{\tau \sim \pi_{\theta_{i}}, s_{0}=s, a_{0}=a}\left[\sum_{t=0}^{\infty} \gamma^{t} r_{t}\right] \\
 Q_{c}^{q}(s, a)=Q_{c}^{\pi_{\theta_{i}}}(s, a)=\mathbb{E}_{\tau \sim \pi_{\theta_{i}}, s_{0}=s, a_{0}=a}\left[\sum_{t=0}^{\infty} \gamma^{t} c_{t}\right]
\end{align}

Considering  \( \mathbb{E}_{q(\cdot \mid s)}\left[Q_{r}^{\pi_{\theta_{i}}}(s, a)\right] \) is on an arbitrary scale, a hard constraint \( \epsilon_{2} \) is introduced on the KL divergence  between  \( q(a|s) \)  and  \( \pi_{\theta_{i}}(a|s) \). Then we obtain:
 \begin{align}
 	\max _{q}~~& \mathbb{E}_{\rho_{q}}\left[\int q(a|s) Q_{r}^{\pi_{\theta_{i}}}(s, a) d a\right] \label{cf1}\\[1pt]
 	\text { s.t. } & \mathbb{E}_{\rho_{q}}\left[\int q(a|s) Q_{c}^{\pi_{\theta_{i}}}(s, a) d a\right] \leq \epsilon_{1} \label{cf2}\\[5pt]
 	 &\mathbb{E}_{\rho_{q}}\left[D_{\mathrm{KL}}\left(q(a|s) \| \pi_{\theta_{i}}\right)\right] \leq \epsilon_{2}  \label{cf3}\\[5pt]
 	 &\int q(a|s) d a=1, \quad \forall s \sim \rho_{q} \label{cf4}
 \end{align}
Where constraint (\ref{cf2}) is to ensure  \( q \)   within the feasible sets,  constraint (\ref{cf3}) is to ensure \( q \) within a trust region of the old policy distribution, and constraint (\ref{cf4}) is to ensure that  \( q \) is a valid action distribution across all the states.

 By the mild assumption (Slator's Condition: $\exists \bar{q}$, \( D_{\mathrm{KL}}\left(\bar{q} \| \pi_{\theta_{i}}\right)<\epsilon_{2}\)), we have  the optimal variational distribution \(q_{i}^{*}(a|s) \) within \( \Pi_{\mathcal{Q}}^{\epsilon_{1}} \) for problem (\ref{cf1})-(\ref{cf4}) has the form:
 \begin{gather}
 q_{i}^{*}(a | s)=\frac{\pi_{\theta_{i}}(a|s)}{Z(s)} \exp \left(\frac{Q_{r}^{\theta_{i}}(s, a)-\lambda^{*} Q_{c}^{\theta_{i}}(s, a)}{\eta^{*}}\right)
 \end{gather} 
 Where \( Z(s) \) is a constant normalizer, and the dual variables \( \eta^{*} \) and \( \lambda^{*} \) are the solutions of convex optimization problem as follows:
 \begin{gather}
 	\min _{\lambda, \eta \geq 0} g(\eta, \lambda)=\lambda \epsilon_{1}+\eta \epsilon_{2}+ \notag\\ 
 	\eta \mathbb{E}_{\rho_{q}}\left[\log \mathbb{E}_{\pi_{\theta_{i}}}\left[\exp \left(\frac{Q_{r}^{\theta_{i}}(s, a)-\lambda Q_{c}^{\theta_{i}}(s, a)}{\eta}\right)\right]\right] \label{gg}
 \end{gather}

From {\bf Theorem 2} in reference~\cite{liu2022constrained},  \( g(\eta, \lambda) \) in  (\ref{gg}) is convex on \( \mathbb{R}_{\geq 0}^{2} \).  Further, strict convexity (in other words, one unique optimal solution) holds if  (1) \( Q_{r}^{\theta_{i}}(s, \cdot), Q_{c}^{\theta_{i}}(s, \cdot) \) are not constant functions; (2) \( \forall C \in \mathbb{R}, \exists a_{0} \), s.t. \( Q_{r}^{\theta_{i}}\left(s, a_{0}\right) \neq \) \( C \cdot Q_{c}^{\theta_{i}}\left(s, a_{0}\right); \) and (3) \( \lambda<+\infty \). 

It is a very nice property that the solution $q$ is always optimal (exact) towards the convex problem, as we strongly desire. This step will output an optimal  $q_{i}^{*}$  for the next M-Step.


 \vskip 0.2cm
 
\subsubsection{M-Step}  The goal is to improve the ELBO (\ref{elbo}) w.r.t $\theta$. By dropping the terms in Eq.(\ref{elbo})  independent of $\theta$, we obtain the objective as follows:
 \begin{align}
\overline{\mathcal{J}}(\theta)=\mathbb{E}_{\rho_{q}}\left[\alpha \mathbb{E}_{q_{i}^{*}(\cdot \mid s)}\left[\log \pi_{\theta}(a | s)\right]\right]+\log p(\theta)
 \end{align}

Like in reference~\cite{abdolmaleki2018maximum}, a Gaussian prior is adopted around the old policy parameter \( \theta_{i} \) in this work: \( \theta \sim \mathcal{N}\left(\theta_{i}, \frac{F_{\theta_{i}}}{\alpha \beta}\right) \), where \( F_{\theta_{i}} \) is the Fisher information matrix and \( \beta \) is a positive constant. With this Gaussian prior, we obtain the generalized objective as follows:
 \begin{align}
\max _{\theta} \mathbb{E}_{\rho_{q}}\left[\mathbb{E}_{q_{i}^{*}(\cdot \mid s)}\left[\log \pi_{\theta}(a | s)\right]-\beta D_{\mathrm{KL}}\left(\pi_{\theta_{i}} \| \pi_{\theta}\right)\right] \end{align}

Similar to the E-step, we convert the soft KL regularizer to a hard KL constraint to deal with different objective scales:
\begin{align}
	\max _{\theta} & \mathbb{E}_{\rho_{q}}\left[\mathbb{E}_{q_{i}^{*}(\cdot \mid s)}\left[\log \pi_{\theta}(a |s)\right]\right] \\
	\text { s.t. } & \mathbb{E}_{\rho_{q}}\left[D_{\mathrm{KL}}\left(\pi_{\theta_{i}}(a | s) \| \pi_{\theta}(a | s)\right)\right] \leq \epsilon
\end{align}

The M-Step is conducted in a supervised learning way to obtain policy $\theta$ and allows the flexible design of data collection.

\section{Preparation and Result}

\subsection{Preparation}

\subsubsection{Vehicle Dynamics, Speed Curve, and Software/Hardware Condition}
\begin{figure}[!tbh]
	\centering
	\includegraphics[width=3in]{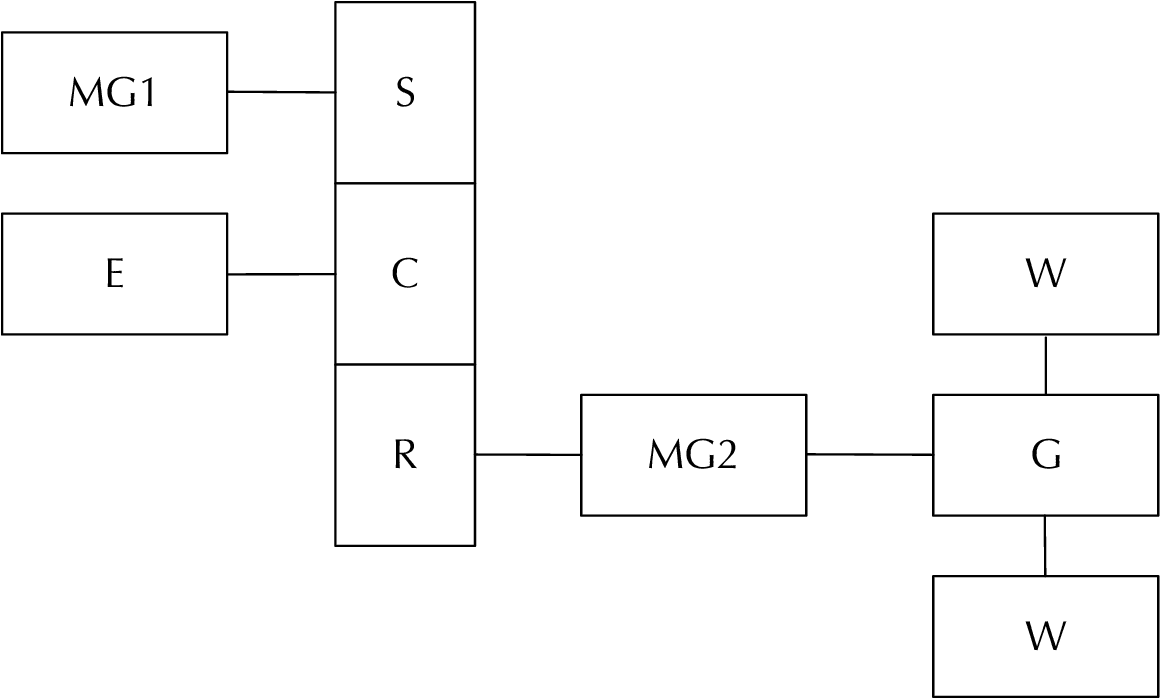}
	\caption{The THS structure.}
	\label{prius1}
\end{figure}

We utilize the well-recognized academic research model  THS  system as in \cite{lian2020rule, liu2023safe}. The structure of THS is shown in Fig.\ref{prius1}. The MG1 and MG2 represent the generator and motor, respectively. E, C, S, R, W, and G represent the engine, carrier gear, sun gear, ring gear, wheel, and main reducing gear, respectively. This vehicle model can be in electric and hybrid modes, and its parameters are the same as \cite{lian2020rule}. For our concern, we need to know the {\sffamily  observation} =\{SOC, velocity, acceleration\}, and the {\sffamily  action}=\{continuous action: engine power\}.   Moreover, in our case, we let THS follow the speed curve of the NEDC condition with a total distance of 10.93km.

We utilize Python 3.9~\cite{van1995python} in PyCharm~\cite{JetBrains} to do a simulation on a MacBook Pro (13-inch, 2017,  2.3 GHz, two Intel Core i5, Intel Iris Plus Graphics 640 1536 MB, 8 GB 2133 MHz LPDDR3, macOS 13.4.1 22F82).

\subsubsection{Procedure}  Currently, most RL approaches are implemented in Python and are interfaced with gym games. Here, we recommend that interested readers write the environment in a gym form through the following steps (assume your environment is named {\sffamily YourEnv}): 

Step 1: Environment Creation: download the gym package, and write the environment as the {\sffamily gym.ENV} form including {\sffamily step}, {\sffamily reset}, {\sffamily render}, {\sffamily close } and {\sffamily  seed}. The {\sffamily step} function is most important to push the dynamic system forward. It is noted that in {\sffamily step} function, the environment can be written in other mainstream languages like Julia~\cite{bezanson2017julia} or Matlab/Simulink~\cite{eshkabilov2019beginning}, and later be called by Python.

\begin{figure}[!tbh]
	\centering
	\includegraphics[width=2in]{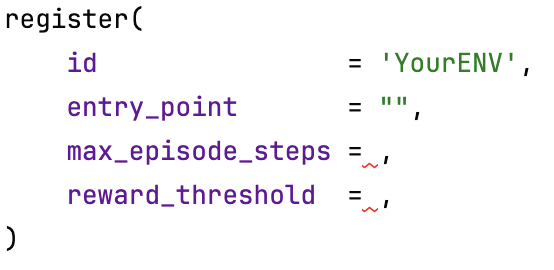}
	\caption{Add 'YourENV' in environment registration.}
	\label{reg}
\end{figure}

Step 2:  Enviroment Registration: write the registration in the {\sffamily \_\_init\_\_.py} file under  {\sffamily /gym/envs/\_\_init\_\_.py} directory as in Fig.\ref{reg}.

Step 3:  Enviroment Import: import {\sffamily YourEnv} in {\sffamily \_\_init\_\_.py } within the same folder by  command {\sffamily from ... import ...}~.

\subsection{Case Study and Results}

\textit {Case Design}: We randomly choose the action within the action limit and get a random SOC trajectory. Then, we choose the SOC-allowed range based on the random SOC trajectory and utilize CVPO and Lagrangian-based approaches to see/check: i) if the CRL approaches can ensure that the SOC balance constraint can be satisfied. ii) the performance difference between CVPO and Lagrangian-based approaches. 

\textit{Parameter Setting and Tuning}:  For the entire simulation parameters, we put it on Google Drive~\cite{crl}. To emphasize the key issues enough, we only give the related parameters in this paper for illustration purposes. For more detailed parameters, we recommend interested readers refer to~\cite{crl}.

For off-policy methods, since \( J_{c}(\pi)=E_{\tau \sim \pi}\left[\sum_{t=0}^{\infty} \gamma^{t} c_{t}\right] \) is the discounted expectation of return of cost, the cost limit $\epsilon_{1}$ can be estimated by 
\begin{gather}
	\epsilon_{1}=\epsilon_{T} \times \frac{1-\gamma^{T}}{T(1-\gamma)}
\end{gather}
Where  $\epsilon_{T}$ is the total cost limit for one episode without considering the discounter factor $\gamma$,  and ${T}$ is the length of one episode. This estimation assumes the same constraint violation for each time step in one episode, and we adopt  $\epsilon_{T}=1.50$ in this work (namely, 	$\epsilon_{1}\approx 0.06$ for state-wise threshold).

This procedure is only needed for off-policy methods because only off-policy methods utilize safety critic (such as $Q_{c}^{\pi_{\theta_{i}}}(s, a)$ in Eq.(\ref{qc}) of CVPO in this paper) as their cost estimations, while on-policy methods do not require it. On-policy methods can directly utilize on-policy rollout cost estimates and the target cost to update the lagrangian multiplier, so there is no need for such a conversion. We start to choose random actions with seed 16 in Python. We started 50 parallel environments, tested the recorded best policy on one environment, and then did 1550-epoch training using CRL approaches, and the default steps per epoch are 3000.

\begin{figure*}
	\flushleft 
	\hspace*{-2.0cm}\includegraphics[width=8.7in]{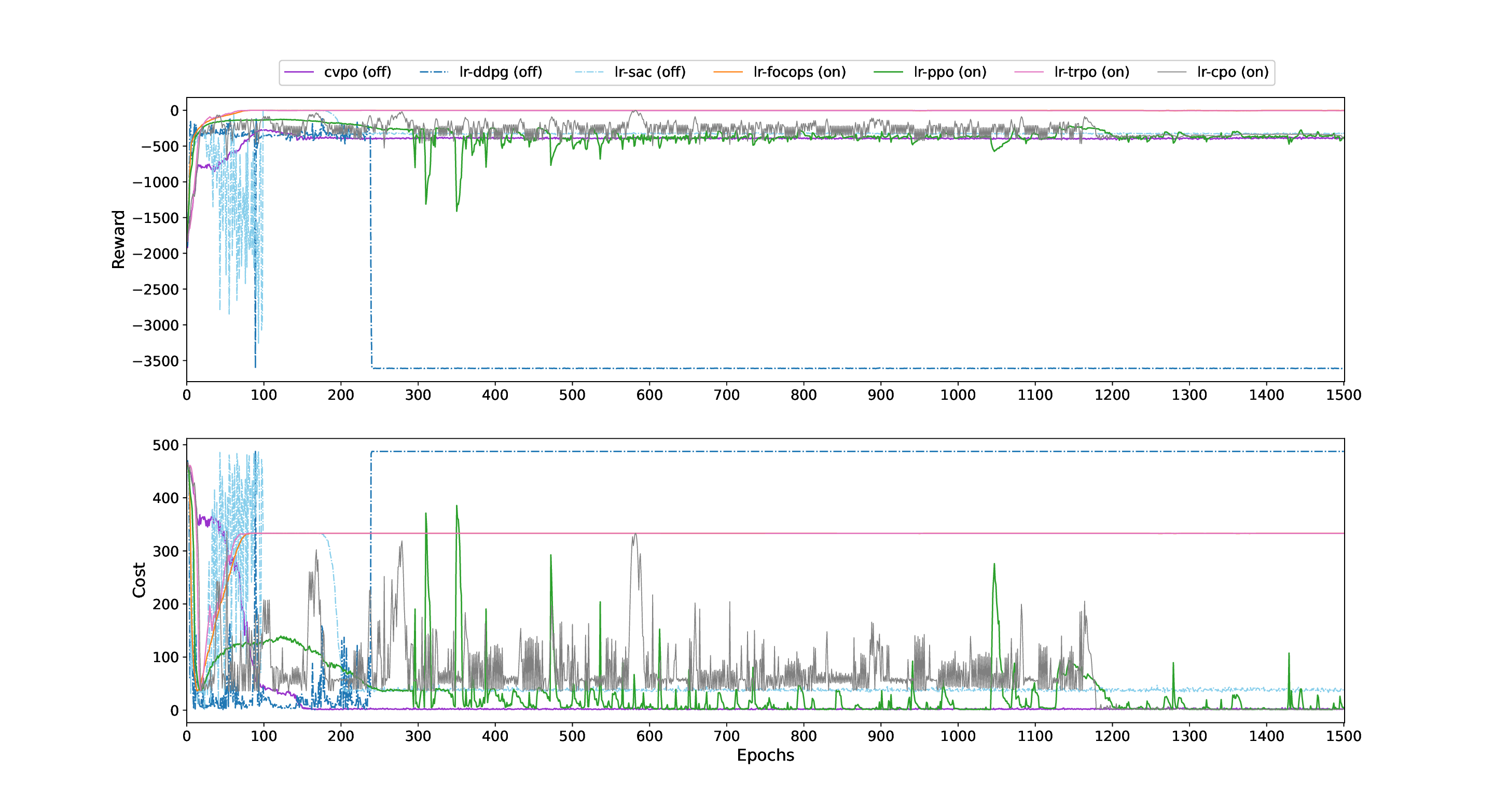}
	\vspace{-1.3cm}
	\caption{Training curves of  CRL approaches. (on) and (off) represents on-policy and off-policy ones, respectively.}
	\label{rc}
\end{figure*}

For the results, Fig.\ref{rc} presents the first 1500-epoch training process, giving the mean training reward and training cost at the subplots above and below. Table~\ref{tab1} gives the best policy's results, including the reward, cost, and training time. 

From Fig.\ref{rc}, we can see: 

i) CVPO has a stable convergence process, and the convergence speed is slower than the Lagrangian ones in the first 100 epochs, and it is almost stable at epoch 162. Finally, the reward converges between -395 and -400, and there are almost no oscillations from the current scale drawing. 

ii) For off-policy Lagrangian approaches (lr-SAC and lr-DDGP), lr-SAC experiences more significant oscillations in the first 100 epochs; the reward converges at around -338, but the cost falls between around 35 to 37.5; lr-DDGP enjoys minor oscillations, and there are many points within the cost limit range 1.5, but finally the training is failed as we can see the reward is around -330 and the cost climbs to around 487. For practical utilization, we can use the first 230 epochs of lr-DDPG to seek the best policy, which has been recorded automatically by our Python code.

iii)   For on-policy Lagrangian approaches (lr-FOCOPS,  lr-PPO,  lr-TRPO, and lr-CPO), interestingly, lr-FOCOPS and lr-TRPO have similar behaviors:  the costs first fall to around 36 at epoch 14 and epoch 18 respectively and finally go up to around 333 respectively. Lr-PPO has a desired training process, as we can see that the cost experiences one oscillation from epoch 1 to epoch 200 and then gradually becomes close to and oscillates around the cost limit of 1.5 and the reward -391. Lr-CPO has a fantastic performance:  it converges much faster than other CRL approaches and is almost stable at around epoch 85 with a reward of around -385 and a cost of around 1.4. For the lr-cpo approach, we are amazed to find that before epoch 1165,  the cost has high-frequency oscillations above about 38, and the reward has high-frequency oscillations around 275. From epoch 1165 to epoch 1179, the cost rapidly falls to around the cost limit of 1.5, and the cost of high-frequency oscillations disappears.

Lagrangian-based approaches may have better training performance with different parameter settings, such as learning rates and discount factor $\gamma$. However, this case study aims to compare the performance of different CRLs and solve the COFC problem, so the values of the identical parameters are set to the same value.

From Table~\ref{tab1}, we can see that the CVPO (off-policy), lr-DDPG (off-policy), lr-PPO  (on-policy), and lr-CPO  (on-policy) approaches can successfully ensure the SOC balance constraint, and others can not. The final SOC of these four approaches is all with $0.50\pm0.03$, and this range definitely can be accepted by the industry. Suppose interested readers want the lower cost and higher reward. In this case, interested readers can increase the epochs for training and make the learning rate finer; this behavior will require more computation time. Under the cost limit 1.50 concern, lr-DDPG (off-policy) gets the best reward (the lowest fuel consumption 311.43  $g$). Moreover, the lr-CPO consumes the most time, more than 36 hours, and the lr-DDPG with the minimum time,  about 6 hours. Another unbelievable thing is that the best policies of lr-FOCOPS (on-policy) and lr-TRPO (on-policy) are almost the same, as we see the SOC dynamics of these two approaches coincide.

\begin{table*}
	\renewcommand{\arraystretch}{1.}
	\caption{Test results recovered by the best policies of CRL approaches.}
	\label{tab1}
	\centering

	\begin{tabular}{lccccccc}
		\hline
		\hline
		Approach  & Type &Reward & Fuel Consumption (L/100km)& Cost & Time (s) &  Final SOC & Performance\\
		\hline
	    Random   & not applicable  &    -1792.76     & 22.76 &     465.62    &  not applicable  &  1.000  & unsatisfactory\\
		\hline
		CVPO         & off-policy &-333.91 		& 4.24 & 	 	1.47 &64616.26  &  0.516 & \color{OliveGreen}{satisfactory}\\
		\hline
		lr-DDPG    &  off-policy&-311.43		& 3.95 &	1.16& 22410.40  &  0.511 & \color{OliveGreen}{satisfactory}\\
		\hline
		lr-SAC        & off-policy  &-805.78	& 10.23 &    355.51		   &   29450.37 &  1.000 &unsatisfactory\\
		\hline 
		lr-FOCOPS & on-policy &0.00	&0.00&	333.04 &41570.36 & -0.530 &unsatisfactory\\
		\hline
		lr-PPO       & on-policy &-338.73 & 4.30 &	1.49 & 48482.46 &  0.511 & \color{OliveGreen}{satisfactory}\\
		\hline
		lr-TRPO     &on-policy & 0.00	& 0.00 & 333.04	&68101.40 & -0.530 & unsatisfactory\\
		\hline
		lr-CPO       &on-policy& -326.76	& 4.15 &	1.09&  131883.89  & 0.525 &  \color{OliveGreen}{satisfactory}\\
		\hline
		\hline
	\end{tabular}

\end{table*}

\begin{figure*}
	\flushleft 
	\hspace*{0cm}\includegraphics[width=7.1in]{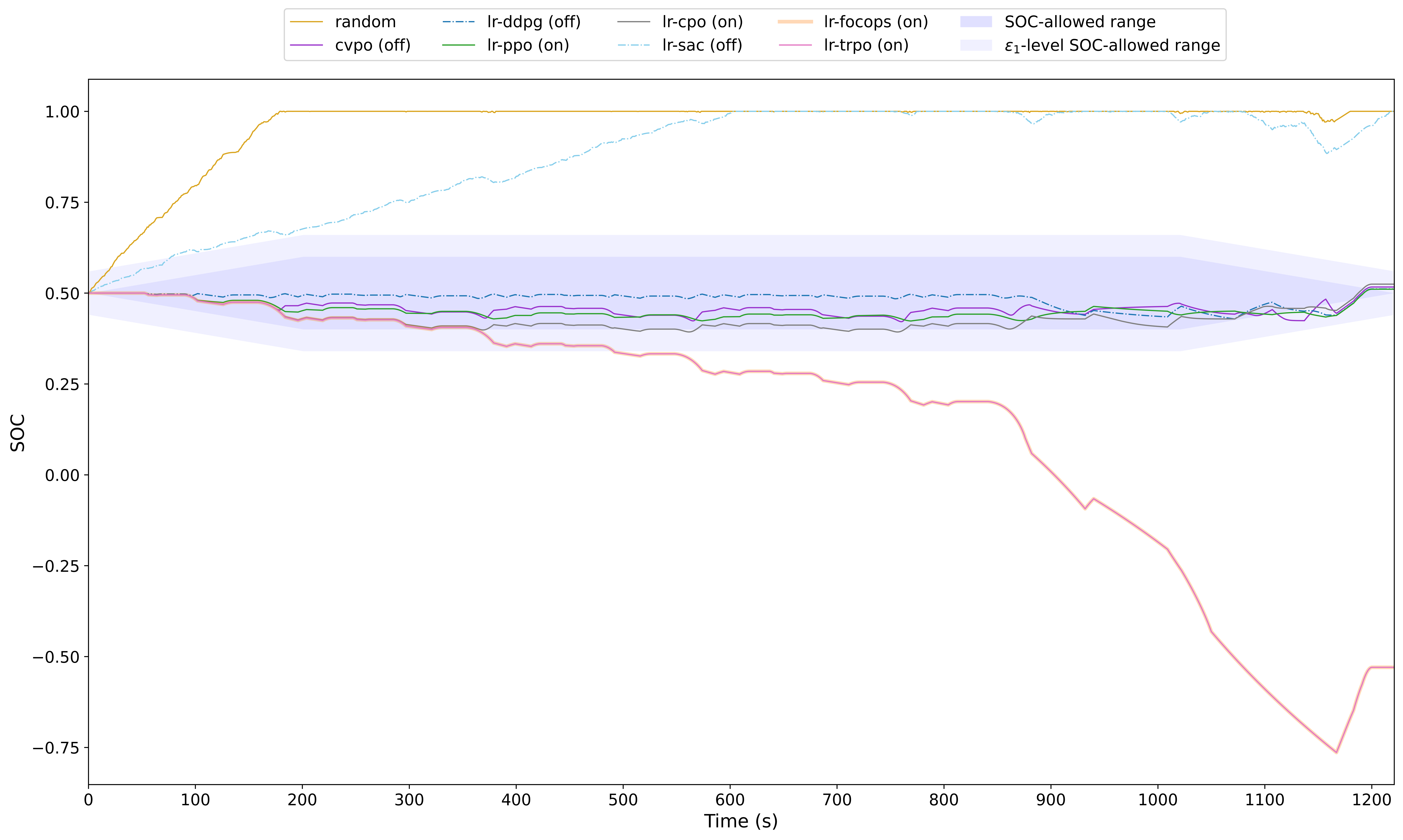}
	\vspace{-0.3cm}
	\caption{The SOC dynamics recovered by the best policies of CRL approaches.}
	\label{soc_region}
\end{figure*}

From Fig.\ref{soc_region}, we can see that SOC trajectories explored by satisfactory CRL approaches are all within the $\epsilon_{1}$-level SOC-allowed range. Moreover, except from around epoch 1110 to 1185, the SOC dynamics are within or slightly out of the SOC-allowed range. This result means the optimal trajectories of system dynamics explored by satisfactory CRL approaches successfully meet the SOC balance constraint and solve the COFC problem.

To reveal more information to readers, we also compare the engine's speed, torque, and working point of our best result from lr-DDPG and our worst result from random action. For other comparisons, interested readers can draw it by themselves through the data we put on the website~\cite{crl}.

\begin{figure*}
	\flushleft 
	\hspace*{-2cm}\includegraphics[width=8.5in]{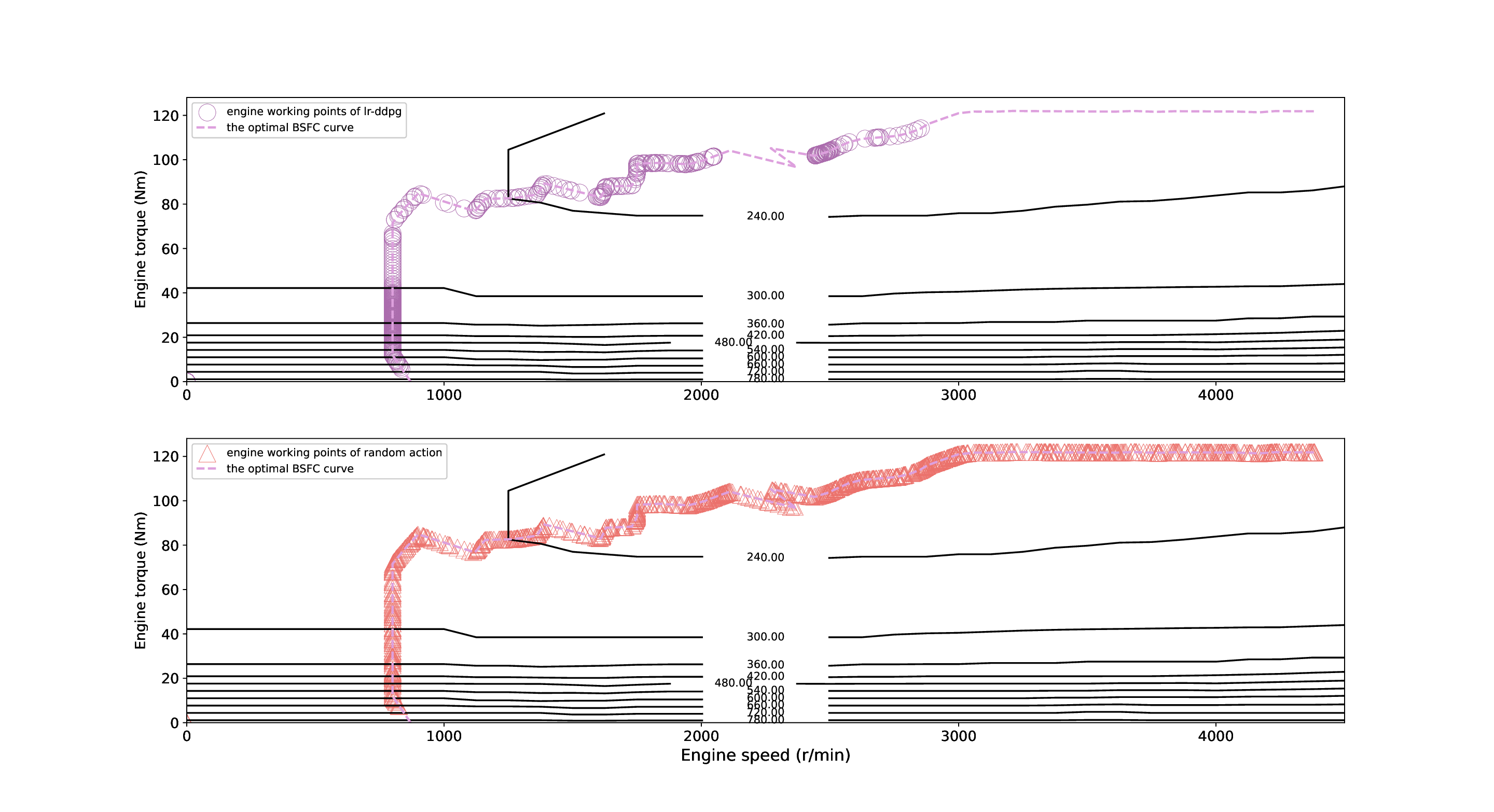}
	\vspace{-0.1cm}
	\caption{Engine working points of lr-DDPG and random action, respectively.}
	\label{wp}
\end{figure*}

\begin{figure*}
	\flushleft 
	\hspace*{-2.0cm}\includegraphics[width=8.5in]{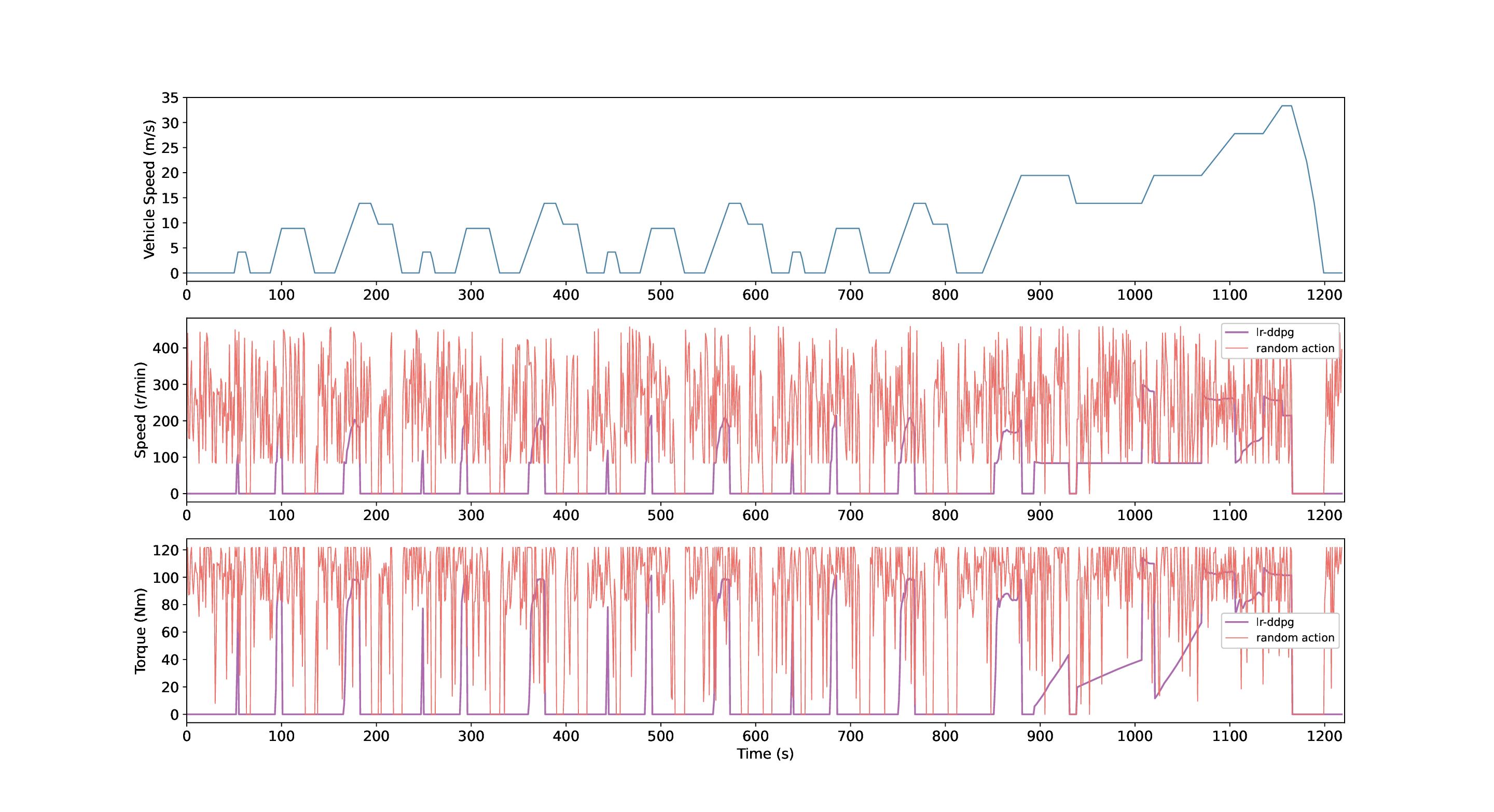}
	\vspace{-0.1cm}
	\caption{Engine speed and torque of lr-DDPG and random action, respectively.}
	\label{ps}
\end{figure*}

From Fig.\ref{wp} and Fig.\ref{ps}, we can see 

i) Random action (or policy) almost randomly scatters the engine power points along the optimal BSFC curve from engine speed around 800 r/min to  4380 r/min. However, the lr-DDPG gathers the engine power points at around 800 r/min and from around 2440 r/min to 2500 r/min, respectively.

ii) Random action (or policy) makes the engine on for most of the NEDC conditions, but the lr-DDPG allows the engine to be on and off to allow the mode change between the HEV and EV.

\section{Conclusion}

In this work, we seminally formulate the COFC problem from the CRL perspective. We perform comparative experiments on random actions and actions generated by the CRL controller. The results show that the CRL approaches can obtain optimal fuel consumption while satisfying the SOC balance constraint. Among these two classes, the CVPO approach can achieve better stability as optimality and feasibility guarantee in its design; Lagrangian-based approaches experience more oscillations, and lr-DDPG obtains the lowest fuel consumption. We confirm that the CRL approaches proposed in this work can solve the burning question raised in section I, which now confuses the auto industry.

This work is the first to answer the question of where the best fuel consumption level is in PEC design. Like a stone thrown on the lake, we expect this paper to impact academia and industry broadly.

\section{Discussions}
The common practice in academia is to utilize research models (like the Prius THS model in our work) to illustrate the effectiveness of methods and better communicate with peers. However, from the perspective of engineering applications, we urgently need to find a fast simulation method for large-scale dynamic systems because the CRL requires fast simulation to collect vast amounts of environmental (vehicle dynamic model) data. Regarding software implementation, Python dominates the programming of CRL, and it will be difficult to change within a period. For environment description, we must find out how fast emulation is for other languages such as Julia. Unfortunately, from our experience, using Simulink to model more complicated vehicle systems requires compilation. Though we can finally get the result, the simulation speed must be faster. Regarding hardware, we know we can speed up simulation by increasing computing power, but the specific implementation method still needs to be determined. 

Speeding up the CRL is also a problem. The CVPO approach must collect multiple particles in the E-step and calculate a convex optimization problem, so the calculation efficiency is low. For Lagrangian-based approaches, we must also answer how to speed up primal-dual iteration. The difficulty lies in calculating the learning rate and Lagrangian multipliers. Could we improve this process to obtain a better speed? Moreover, no existing CRL approaches can guarantee the solution's optimality, so a new perspective we can think of is to know under what conditions we can relax the problem.

This work has paved the way for a new stream of work on the COFC problem, which needs more talent to explore. We are also willing to discuss technical issues with researchers and engineers to contribute to the PEC design and the COFC problem.

\section*{Acknowledgments}
We thank Dr. Zuxin Liu at Carnegie Mellon University, who proposed the CVPO approach, for his advice on our work in early 2024. We have discussed the good points and shortcomings of the CVPO approach and clarified a few mathematical descriptions around the CVPO approach together.






\ifCLASSOPTIONcaptionsoff
\newpage
\fi

\bibliographystyle{IEEEtran}
\bibliography{journal}


\vspace{11pt}





\vfill

\end{document}